\documentclass[conference,compsoc,11pt]{IEEEtran}
\IEEEoverridecommandlockouts

\ifCLASSOPTIONcompsoc
  \usepackage[nocompress]{cite}
\else
  \usepackage{cite}
\fi

\usepackage{amsmath,amssymb,amsfonts}
\usepackage{algorithmic}
\usepackage{graphicx}
\usepackage{textcomp}
\usepackage{xcolor}
\usepackage{array}   %
\usepackage{tabularx} %
\usepackage{graphics} %
\usepackage{multirow}
\usepackage{booktabs}

\usepackage{float}

\usepackage{siunitx}

\ifCLASSINFOpdf
\else
\fi

\begin{document}
\title{Methodology for an Analysis of Influencing Factors on 3D Object Detection Performance
\\
\thanks{The research is accomplished within the project “AUTOtech.agil” (FKZ 01IS22088S). We acknowledge the financial support for the project by the Federal Ministry of Education
and Research of Germany (BMBF)}
}

\author{\IEEEauthorblockN{Anton Kuznietsov}
\IEEEauthorblockA{\textit{Institute of Automotive Engineering} \\
\textit{Technical University of Darmstadt}\\
Darmstadt, Germany  \\
anton.kuznietsov@tu-darmstadt.de}
\and
\IEEEauthorblockN{Dirk Schweickard}
\IEEEauthorblockA{\IEEEauthorblockA{\textit{Institute of Automotive Engineering} \\
\textit{Technical University of Darmstadt}\\
Darmstadt, Germany  \\
dirk.schweickard@online.de}}
\and
\IEEEauthorblockN{Steven Peters}
\IEEEauthorblockA{\textit{Institute of Automotive Engineering} \\
\textit{Technical University of Darmstadt}\\
Darmstadt, Germany \\
steven.peters@tu-darmstadt.de} }

\maketitle

\maketitle

\begin{abstract}
In automated driving, object detection is crucial for perceiving the environment. Although deep learning-based detectors offer high performance, their black-box nature complicates safety assurance. We propose a novel methodology to analyze how object- and environment-related factors affect LiDAR- and camera-based 3D object detectors. A statistical univariate analysis relates each factor to pedestrian detection errors. Additionally, a Random Forest (RF) model predicts errors from meta-information, with Shapley Values interpreting feature importance. By capturing feature dependencies, the RF enables a nuanced analysis of detection errors. Understanding these factors reveals detector performance gaps and supports safer object detection system development.
\end{abstract}

\section{Introduction}
The foundation of perception in automated driving (AD) is object detection by various sensors, including RGB cameras, LiDAR, and radar \cite{ignatious2022overview}. To realize the potential of automated vehicles (AVs), it is necessary, among other factors, that AVs perceive their environment accurately and robustly \cite{sun2023toward}. High performance in object localization and classification is achieved by object detectors based on deep neural networks (DNN) \cite{mao20233d}. However, DNNs exhibit several safety concerns such as brittleness or sensitivity to small perturbations, and unknown behavior due to their black box nature \cite{Willers2020}. Thus, providing safety assurance for DNNs is challenging \cite{burton2022safety}, which is a major problem for the applicability and certification of DNN-based object detectors in the safety-relevant domain of AD. A key step in the Machine Learning (ML) safety lifecycle for the development and deployment of ML-based models is the evaluation of performance insufficiencies \cite{Burton2023Uncertainty}. In contrast to previous studies that primarily identify sensor-specific hazards or investigate individual influencing factors on perception algorithms, this work systematically identifies and analyzes the impact of various meta-information on the overall detection performance of DNN-based 3D object detectors across different modalities. Meta-information refers to the characteristics of the scene, such as weather conditions or specific object properties. Since the impact of influencing factors varies across object classes due to their different characteristics, we focus on pedestrians, one of the most safety-relevant object classes. By analyzing camera-based and LiDAR-based single-sensor 3D object detectors, we also investigate the differences between the two sensor modalities. Our developed methodology identifies specific weaknesses for the different object detectors and supports the safety approval process of DNN-based object detectors.\\
The main contributions of our work are: (i) providing a comparison of detectors' dependencies on meta-information, (ii) identifying the differences in the influencing factors between LiDAR-based and camera-based single-sensor detectors, and (iii) identifying individual dependencies of different detectors.

\section{Related Work}
While there are studies that analyze the hazards within the perception chain, most of them focus on the hazards resulting from specific sensor properties or only examine specific effects on perception algorithms. However, a comprehensive analysis that takes into account the various factors influencing 3D object detection is still lacking. Zendel et al. \cite{Zendel.2015} present CV-HAZOP, which identifies multiple hazards for camera-based computer vision systems. They apply a Hazard and Operability analysis (HAZOP) to a generic camera-based computer vision model. After computer vision experts identify and specify hazards, the corresponding hazards are manually annotated in public datasets. Subsequently, several conventional stereo vision algorithms are experimentally applied to the test data and evaluated based on misdetected pixels, compared to the ground truth in the image. In \cite{Zendel.2023}, Zendel et al. extended their work by applying a HAZOP to LiDAR sensors resulting in a joint hazard analysis for the camera and LiDAR. A hazard detector is applied to the camera images of the A2D2 dataset \cite{A2D22020} and experts evaluate the quality of the corresponding LiDAR data based on these hazards. This identifies joint hazards for camera and LiDAR, such as reflectivity and transparency properties of objects. Linnhoff et al. \cite{Linnhoff.2021} proposed a Perception Sensor Collaborative Effect and Cause Tree (PerCoLLECT) where cause and effect chains for camera, radar, and LiDAR are developed. Their work covers the entire perception chain, spanning from emission to object identification. An explicit quantitative analysis of hazards on deep learning-based object detectors is not performed in these works. \\
In \cite{Li.05.08.2023}, Li et al. performed a statistical analysis of deep learning-based 2D camera pedestrian detection to evaluate fairness by examining the detector performance only over human attributes like age and gender. 
Finally, Ponn et al. \cite{Ponn.2020} were the first to perform a deep learning-based object detection performance analysis according to the meta-information contained in the object or environment by training a RF \cite{ho1995random}. Subsequently, they applied SHapley Additive Explanations \cite{Lundberg2017SHAP} to rank the importance of each feature respectively meta-information. Although this meta-model-based analysis accounts for multivariate dependencies between features, the learned relationship by the ML-based meta-model does not necessarily represent the true nature of the problem \cite{freiesleben2024supervised}. Furthermore, their work focuses only on the performance of 2D camera-based detectors and only analyzes False Negative (FN) errors, i.e. when an object was not detected. False Positive (FP) errors, i.e. ghost detections, were not examined. \\
The thesis of Liang \cite{Liang2022FailureCause} tries to derive the causes of errors in 3D object detection. However, the analysis also focuses exclusively on FN errors and only examines one camera and one LiDAR-based detector, limiting comparability. Furthermore, the choice of the meta-information and the proposed univariate analysis proved to be insufficient to achieve the goal of accurately identifying error causes. \\
In contrast to previous works, we propose a methodology to analyze the influence of a broader collection of meta-information on the performance of 3D pedestrian detection by performing both univariate and meta-model-based analysis. We also investigate multiple camera- and LiDAR-based detectors, as well as FN and FP errors. 

\section{Methodology}
This section outlines the methodology for analyzing factors affecting detection performance, including meta-information, selected object detectors, and our analytical approach.
\subsection{Meta-information}
Our analysis requires a dataset containing LiDAR and camera sensors with corresponding 3D bounding box annotations and a high variability in scenarios. As in the work of Ponn et al., \cite{Ponn.2020}, we choose the multi-modal and open-source dataset NuScenes \cite{Caesar.2020}. NuScenes has a large amount of data with $40k$ annotated frames that were recorded during the day and at night as well as for different weather conditions. Moreover, it comprises different types of sensors, including 6 cameras and a LiDAR, which together provide a $360^\circ$ sensor coverage. \\
To investigate the influence of meta-information, we extract it from the environmental and ground-truth object annotations contained in NuScenes. We aim to extract a wide range of meta-information in order to identify as many relevant influencing factors as possible. For the environmental meta-information, we examine location, which includes one district in Boston and three districts in Singapore. We consider the time of day and the month of the recording. Next, we consider the presence of rain, as annotated by NuScenes for each sample. Due to the limited environmental information provided by NuScenes, we also extract weather data recorded by the weather station closest to the global ego-position at the time of recording from Visual Crossing's weather API \cite{VisualCrossing}. This includes the temperature and the dew point temperature in $\si{\degreeCelsius}$, the humidity in percent, the precipitation in $\si{\frac{\milli\meter}{\hour}}$, the windspeed in $\si{\frac{\kilo\meter}{\hour}}$, the relative direction from which the wind is blowing, the sea level pressure in $\si{\hecto\pascal}$, the percentage amount of cloud cover, the solar-radiation in $\si{\frac{\watt}{\meter\squared}}$, the visibility in $\si{\kilo\meter}$ and finally the UV Index.  
The meta-information associated with each detected pedestrian comprises its attribute and its category. The attribute specifies the pedestrian's posture, i.e. whether the pedestrian is standing, moving, sitting, or lying down, while the category classifies the pedestrian as a police officer, worker, child, or adult. Moreover, we consider the distance between the ego-vehicle and the pedestrian in $\si{\metre}$, the velocity of the object in $\si{\frac{\metre}{\s}}$, and the relative heading respectively the yaw angle of the object in $\si{\degree}$. Additionally, the width, height, and length of the 3D bounding box around the pedestrian $\si{\metre}^3$ are included. Ponn et al. \cite{Ponn.2020} also investigate the influence of the number of pixels in the camera image. As our analysis includes LiDAR-based object detectors, this factor is not a suitable variable to compare the performance between LiDAR- and camera-based detectors. Therefore, we consider the vertical and horizontal angular size of an object, which represents the apparent size of an object from the ego vehicle's perspective. The angular size is influenced by the object's physical dimensions and its distance to the sensor. Specifically, a larger object and a shorter distance result in a greater angular size, whereas a smaller object and a greater distance lead to a reduced angular size. Finally, we analyze the occlusion of the object, represented by the visibility token provided by NuScenes for each object. The visibility token is categorized into four bins, indicating the extent to which the ground-truth bounding box annotation is visible across all images in the sample. Higher bin values correspond to lower levels of occlusion, while lower bin values represent higher levels of occlusion.   
For FN errors, meta-information about the pedestrian is captured by the ground-truth annotations provided by NuScenes. For FP errors, only object-specific meta-information derived from the predicted bounding box is available. Consequently, the meta-information is limited to the distance, size, and orientation of the predicted bounding box generated by the detector.   
\subsection{Object Detectors}
For single camera-based object detection, we investigate two monocular-based detectors that take one camera image as an input. This type of detector combines 2D detection frameworks with geometric priors to estimate 3D bounding boxes \cite{Chen.2023}. In our study, we analyze the effect of the meta-information on the one-stage detector FCOS3D \cite{Wang.2021} and on the two-stage detector, MonoDIS \cite{Simonelli.2022}. We also analyze the performance of the multiview camera detector SpatialDETR \cite{DollSpatialDETR2022}, a transformer-based architecture. Transformers use self-attention mechanisms to model global relationships between image features and object queries. \\
For LiDAR-based object detection, we analyze two voxel-based object detectors, namely Megvii \cite{Zhu.26.08.2019} and PointPillars \cite{Lang.2019}. Both detectors convert the point cloud into 3D voxels, thereby extracting features for 3D object detection. Finally, the transformer-based LiDAR detector TransFusion-L \cite{Bai2022TransFusion} is also considered. The selection of object detectors was made based on the importance of these detectors in the field of single-sensor detection. Moreover, the architectures are diverse and the two transformer-based detectors represent modern state-of-the-art architectures. \\
Inspired by the AP-score calculation in NuScenes, a detection is considered correct if the object classification is true and the center distance between the predicted and ground-truth locations in Bird's Eye View is less than $2\si{\metre}$. In addition, the confidence thresholds of each detector are optimized based on the F1-score evaluated on the NuScenes dataset. Each detector assigns a confidence score to its predictions, with those below the threshold being discarded. Table \ref{tbl:recall} presents the recall and precision values for pedestrian detection, along with the corresponding optimized confidence thresholds. 
\begin{table}[!ht]
\caption{The detectors with their corresponding sensing modality (c = Camera, L = LiDAR), optimized confidence threshold (Conf. th.), and the Recall as well as Precision for pedestrian detection on the NuScenes validation dataset}
\label{tbl:recall}
\begin{center}
\begin{tabular}{|c|c|c|c|c|}
\hline
Detector & Modality & Conf. th.  & Recall & Precision \\
\hline
FCOS3D & C & 0.2 &  0.57 & 0.69 
\\
\hline
MonoDis & C & 0.2 & 0.52 & 0.73 
\\
\hline
SpatialDETR & C & 0.36 & 0.53 & 0.72 
\\
\hline
Megvii & L & 0.27 & 0.72 & 0.88 
\\
\hline
PointPillars & L & 0.25 & 0.63 & 0.71 
\\
\hline
TransFusion-L & L  & 0.14 & 0.85 & 0.82 
\\
\hline
\end{tabular}
\end{center}
\end{table} 
\subsection{Analysis}
In this study, we aim to compare the influence of meta-information on object detection performance by ranking its influence. We first perform a univariate statistical analysis and then a meta-model-based analysis. For the univariate analysis, we examine Kendall's Tau rank correlation \cite{KENDALL1938} because it has the advantage of being independent of the data distribution. The correlation coefficient expresses negative or positive monotonicity between two variables and ranges between -1 and 1. Its value is $0$ if the variables are independent. The categorical meta-information, including pedestrian attributes, category, and recording location, is encoded using one-hot encoding. Each category is represented as a separate binary variable, assigned a value of 1 if present and 0 otherwise. The inability to capture nonlinear and non-monotonic dependencies between two variables is a shortcoming of Kendall's Tau. Therefore, we additionally use mutual information (MI), which measures the statistical dependence between two variables \cite{Shannon.1948}. MI, which is derived from the union of the conditional entropies of two random variables, effectively captures both linear and nonlinear relationships between two variables, regardless of their monotonicity. High MI values between the detection error and any given factor indicate strong impact, whereas low MI values value reflect weak impact, with the minimum MI value of $0$ indicating no dependency. In addition to its ability to capture nonlinearities, MI can also handle mixed data types by using the k-nearest neighbor entropy estimator for numerical variables \cite{Kozachenk0.1987}. In \cite{Ross.2014} and \cite{Kraskov.2004}, it has been demonstrated that the nearest neighbor approach has lower systematic errors compared to the discretization of numerical variables by binning. As suggested by Kraskov et al. \cite{Kraskov.2004}, we choose a small value of $k=3$, which indicates negligible random error and low systematic error. Furthermore, we normalize the MI value by summing up the individual entropies of the two random variables \cite{Kvalseth1987NormMI}. The lack of consideration of interactions between features is a drawback of univariate analysis. Therefore, we additionally perform a meta-model-based analysis by training an RF \cite{ho1995random} as in the work of Ponn et al. \cite{Ponn.2020}. The RF learns a mapping from the meta-information to the outcome by considering the relationships between the features. For the categorical meta-information, a generalized linear mixed model (glmm) encoding is applied, which shows superior performance in ML-based prediction tasks \cite{glmmPargent2022}. We perform a grid search for the hyperparameter tuning and optimize towards the F1-score. Afterwards, we compare the importance of the features by applying SHapley Additive Explanations (SHAP) \cite{Lundberg2017SHAP}. To ensure that the importance of features reflects the underlying characteristics of the data rather than just the behavior of the model, we use conditional sampling in the computation of Shapley scores to prevent extrapolation of correlated features \cite{freiesleben2024supervised}. The limitation of the model-specific approach is that the trained RF does not necessarily represent the intrinsic characteristics of the problem. By comparing the results with the univariate analysis and considering correlations between the meta-information, we improve the reliability of the analysis.   
We distinguish between FP and FN errors by performing 2 different analyses. 
\section{Results}
This section presents the results of our analysis of influencing factors. First of all, the analysis of the detectors' performance in terms of FN errors is presented. The analysis examines whether an object is detected or not. Then, the analysis of FP errors is then presented, which examines whether a detection generated by the object detector corresponds to a true object or a ghost detection. Both analyses are based on the NuScenes validation dataset.  
\subsection{False Negative Errors}
\begin{figure*}[!ht]
  \centering
    \includegraphics[width=0.75\linewidth]{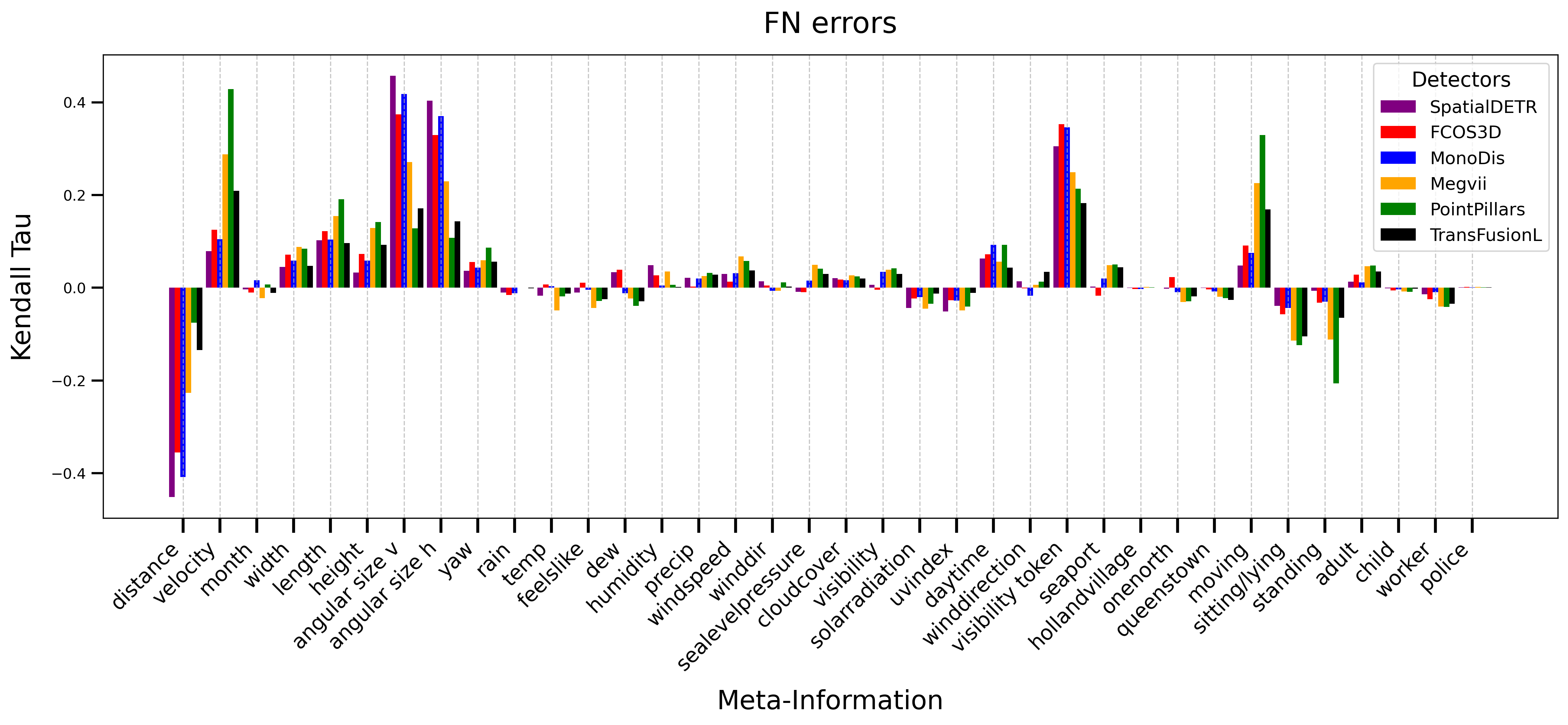}
   \caption{A barplot of the Kendall's Tau rank coefficient between the FN object detection error on pedestrians and the meta-information}
   \label{fig:Kendall_TPFN}
\end{figure*}

\begin{figure*}[!ht]
  \centering
  \begin{minipage}[b]{0.4\linewidth}
    \centering
    \includegraphics[width=\linewidth]{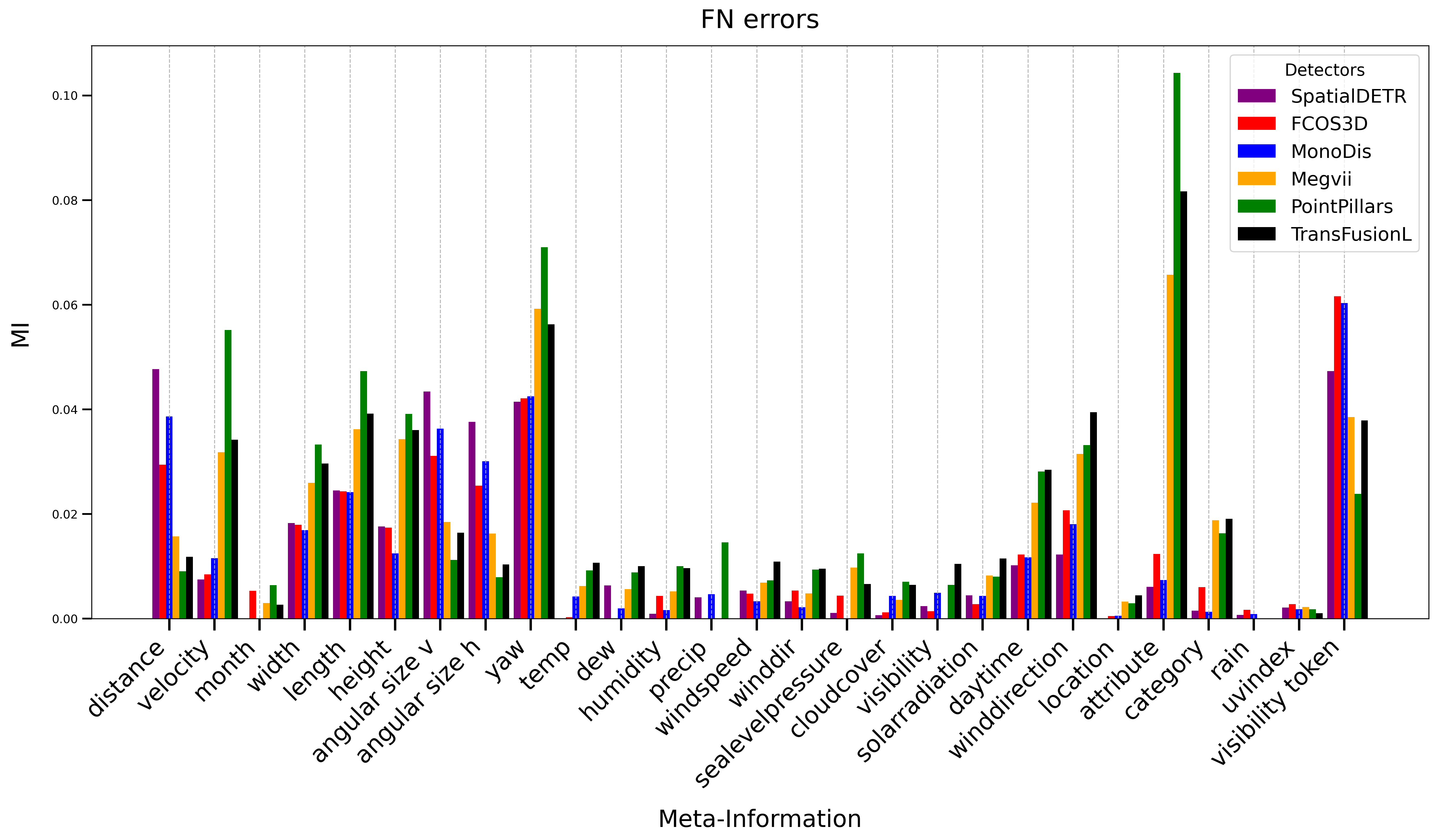}
   \caption{A barplot of the MI value between the FN object detection error on pedestrians and the meta-information}
   \label{fig:MI_TPFN}
  \end{minipage}
  \hfill
  \begin{minipage}[b]{0.45\linewidth}
    \centering
    \includegraphics[width=\linewidth]{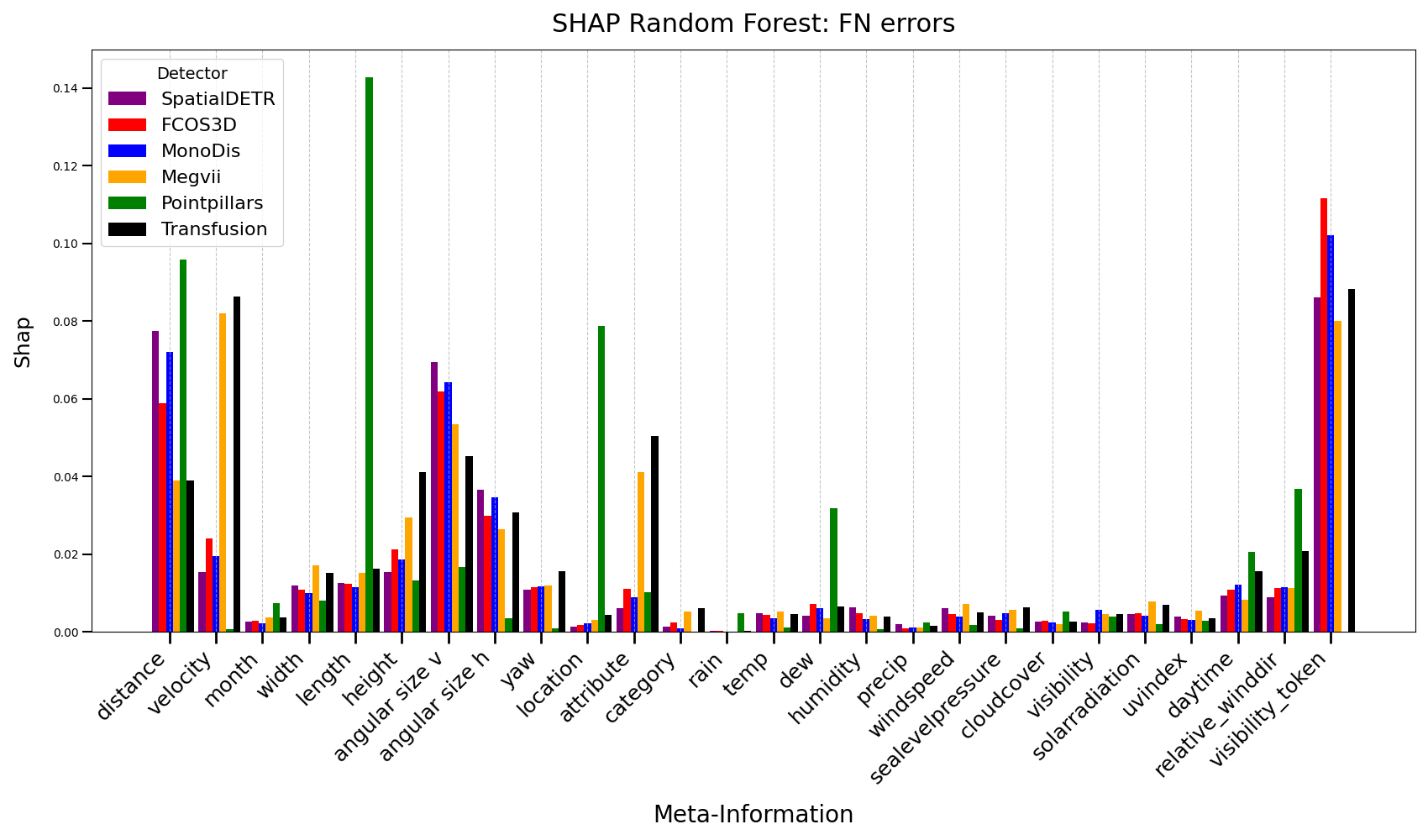}
   \caption{A barplot of the mean absolute shap values between the FN object detection error on pedestrians and the meta-information}
   \label{fig:Shap_FN}
  \end{minipage}
\end{figure*}

First, the FN errors are examined. As shown in Table \ref{tbl:recall}, the recall values for pedestrian detection among the examined detectors indicate that LiDAR-based models, particularly TransFusion-L, outperform camera-based detectors in detecting existing pedestrians. Fig. \ref{fig:Kendall_TPFN} shows a barplot of the Kendall's Tau correlation coefficient between the meta-information, including object-specific ground-truth information and environmental conditions, and whether the object detector detects the pedestrian correctly or not at all. Each bar color corresponds to a particular detector. The purple, red, and blue bars represent the three camera detectors (SpatialDETR, FCOS3D, MonoDIS), while the orange, green, and black bars represent the three LiDAR detectors (Megvii, PointPillars, TransFusion-L). The x-axis shows the meta-information and the y-axis shows the correlation coefficient. A positive value indicates a positive correlation between higher meta-information values and TPs, and a negative correlation with FNs. Conversely, a negative value indicates a negative correlation between higher meta-information values and TPs, and a positive correlation with FNs. As mentioned before, for Kendall's Tau analysis, the attributes of the pedestrian (moving, sitting/lying, standing), the location (boston-seaport, singapore-hollandvillage, -onenorth, -queenstown), and the category (adult, child, worker, police) are one-hot encoded. 
In Fig. \ref{fig:Kendall_TPFN}, it can be observed that camera detectors have a relatively strong negative correlation with distance compared to LiDAR detectors, indicating that more distant pedestrians are more difficult to detect for camera-based systems than for LiDAR-based systems. This trend is further reflected in the positive correlation between camera-based detections and angular sizes that are affected by distance. In particular, shorter distances result in larger angular sizes. In contrast, the LiDAR-based detectors show higher correlation coefficients with pedestrian attributes. In particular, there is a positive correlation between TP and moving pedestrians, while a negative correlation is observed for standing, sitting, or lying pedestrians. This is further evidenced by the strong positive correlation coefficient between the TPs of the LiDAR-based detectors and the velocity of the pedestrians. Furthermore, all detectors perform better when the object has a higher visibility token, i.e. less occlusion of the pedestrian to be detected, especially the camera-based detectors where the coefficient is slightly larger. The correlation coefficients between the camera-based detectors show only small deviations. In contrast, larger differences are observed between the LiDAR-based detectors. Specifically, PointPillars shows a stronger positive correlation with moving pedestrians and velocity, and a stronger negative correlation with standing pedestrians compared to the other detectors. This suggests that PointPillars is more sensitive to pedestrian attributes. In addition, Megvii has a greater negative correlation with distance and a stronger positive correlation with angular size compared to the other LiDAR detectors. This indicates that Megvii has more difficulty detecting objects at longer distances compared to PointPillars and TransFusion-L.
Finally, it can be observed that the meta-information related to weather, time, and location does not show a strong monotonic correlation with detection performance. Fig. \ref{fig:MI_TPFN} shows a barplot of the MI-value between meta-information and detection performance in terms of FN errors. Unlike Kendall's Tau, MI can capture nonlinear non-monotonic dependencies, as it quantifies the strength of dependencies without assigning a directional influence. A similar trend as observed in the Kendall's Tau analysis in Fig. \ref{fig:Kendall_TPFN} is reflected in the MI analysis in Fig. \ref{fig:MI_TPFN}. Camera-based detectors show a stronger dependence on distance and angular size, while LiDAR-based detectors show a stronger dependence on velocity and pedestrian-related attributes.
In particular, the attribute has a strong influence on the LiDAR-based detection performance, in contrast to the other investigated factors. It is also evident that all detectors, especially the camera-based ones, are strongly affected by occlusion. In contrast to the Kendall's Tau correlation coefficient, the MI-values for the height, width, length, and especially the yaw angle of the bounding box are relatively high. This difference between Kendall's Tau and MI suggests that there is a nonlinear, non-monotonic dependence between the detector performance and the bounding box size and orientation. However, the LiDAR detectors are more sensitive to the bounding box parameter than the camera-based detectors. They are also more affected by variations in daytime and wind direction. Other than that, the factors related to weather and location do not have a large impact on performance. \\ Differences between the LiDAR-based detector are also observed. As observed in the Kendall's Tau analysis in Fig. \ref{fig:Kendall_TPFN}, the MI-analysis in Fig. \ref{fig:MI_TPFN} shows that PointPillars is more sensitive to pedestrian attributes and velocity. In contrast, PointPillars appears to be less sensitive to the visibility token respectively occlusion compared to Megvii and Transfusion-L, a behavior not observed before in the Kendall's Tau analysis in Fig. \ref{fig:Kendall_TPFN}. Following the univariate analysis, Fig. \ref{fig:Shap_FN} shows the results of the meta-model-based analysis. As previously described, an RF model is trained to predict whether an existing pedestrian is detected (TP) or not (FN) based on meta-information. In order to quantify the importance of each meta-information variable, the mean absolute SHAP values are computed to assess their impact on the RF model's prediction. Consistent with the results from Kendall's Tau and MI analysis, the mean absolute SHAP values in Fig. \ref{fig:Shap_FN} indicate that distance, angular sizes, and the visibility token or the occlusion of the object have the strongest influence on the performance of the camera-based detectors. However, the vertical angular size has a greater effect than the horizontal angular size. While there are minor differences between the camera-based detectors, their general trends remain similar. In contrast, the influence of meta-information on LiDAR detectors is very different from the results of the univariate analysis in Fig. \ref{fig:Kendall_TPFN} and Fig. \ref{fig:MI_TPFN}. Megvii and Transfusion-L show a similar behavior, expressed by a strong influence of occlusion, velocity, and the pedestrian's attribute. These patterns are also observed in the MI-analysis in Fig. \ref{fig:MI_TPFN}. However, unlike in the MI analysis, angular sizes play an important role in both detectors' performance, as observed in Fig. \ref{fig:Shap_FN}. PointPillars, on the other hand, behaves quite differently from Megvii and Transfusion-L. While the ground-truth bounding box parameters generally have a low mean absolute SHAP value for all detectors, pedestrian length emerges as the most influential factor for PointPillars. Moreover, the key influencing factors of PointPillars, expressed by the mean absolute SHAP values in Fig. \ref{fig:Shap_FN}, also differ strongly from the previous MI-analysis in Fig. \ref{fig:MI_TPFN}. The attribute, the velocity, and the occlusion, which have high MI-values, have relatively low SHAP values for PointPillars. Instead, in addition to the length of the pedestrian, the distance and the location have a large impact on the detection performance. In addition, relative wind direction and dew point temperature, factors not identified as influential in previous analyses, have a moderate influence on the detection performance of PointPillars. 

\subsection{False Positive Error}
\begin{figure*}[htbp]
  \centering
    \includegraphics[width=0.75\linewidth]{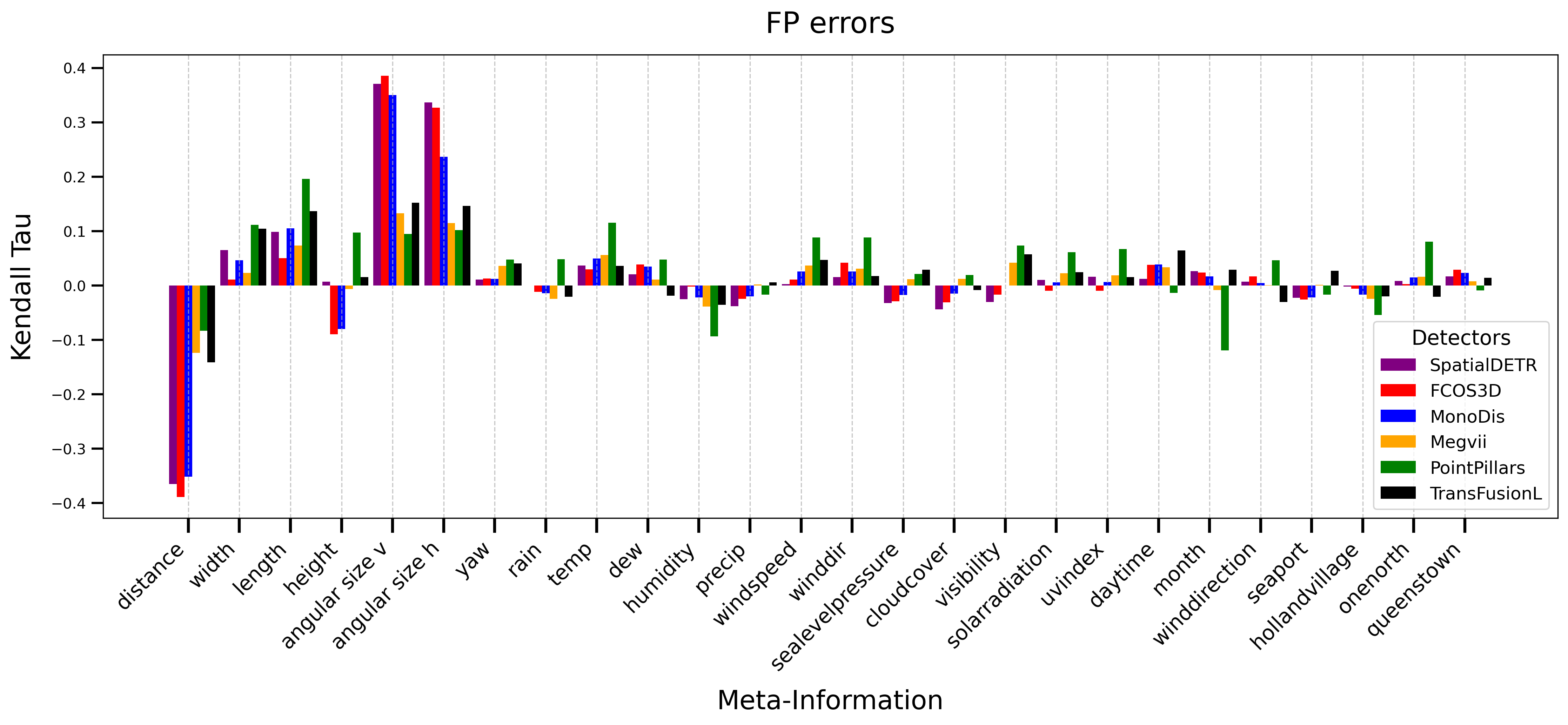}
   \caption{A barplot of the Kendall's Tau rank coefficient between the FP object detection error on pedestrians and the meta-information}
   \label{fig:Kendall_TPFP}
\end{figure*}

\begin{figure*}[htbp]
  \centering
  \begin{minipage}[b]{0.45\linewidth}
    \centering
    \includegraphics[width=\linewidth]{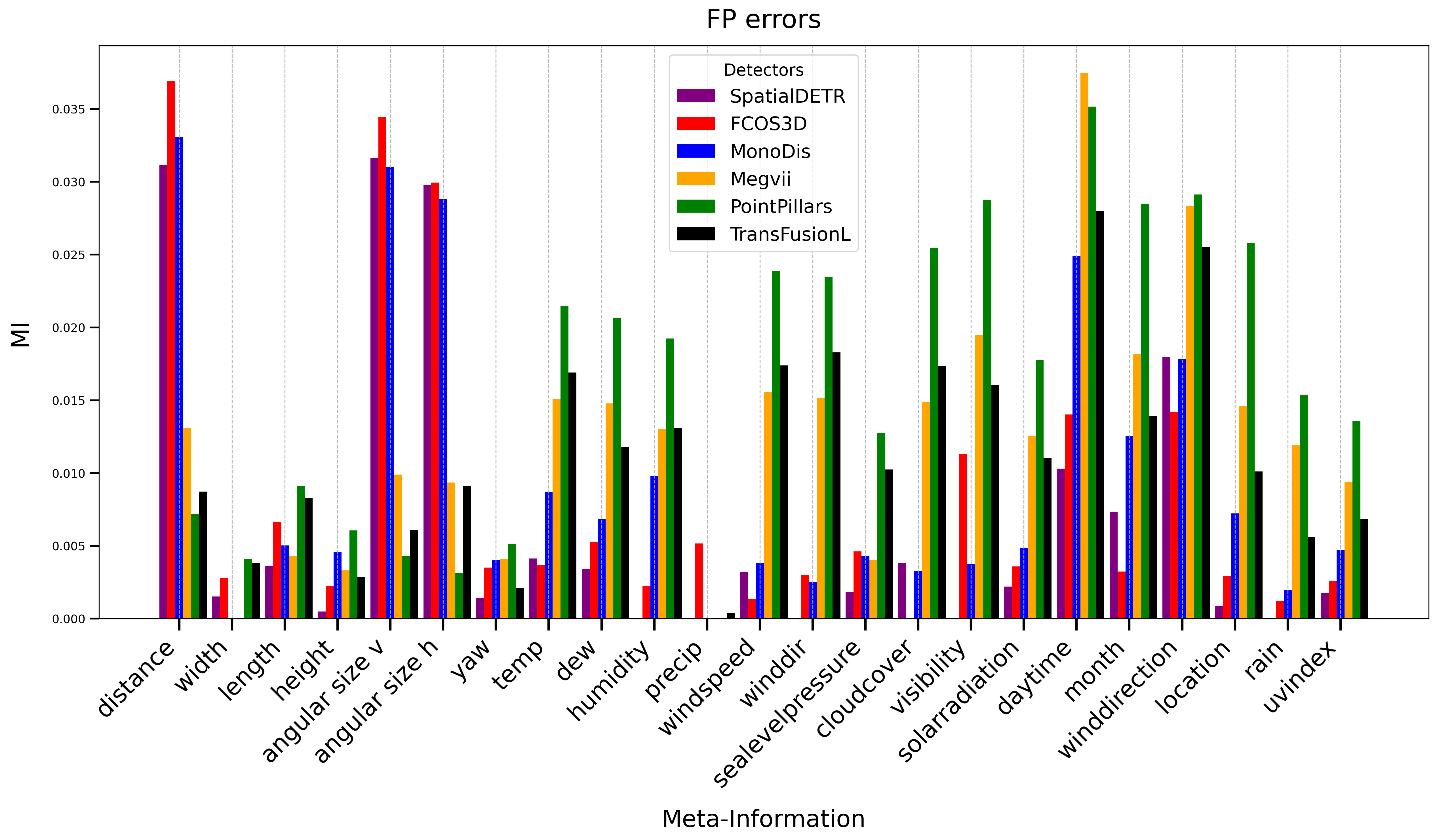}
   \caption{A barplot of the MI value between the FP object detection error on pedestrians and the meta-information}
   \label{fig:MI_FPTP}
  \end{minipage}
  \hfill
  \begin{minipage}[b]{0.45\linewidth}
    \centering
    \includegraphics[width=\linewidth]{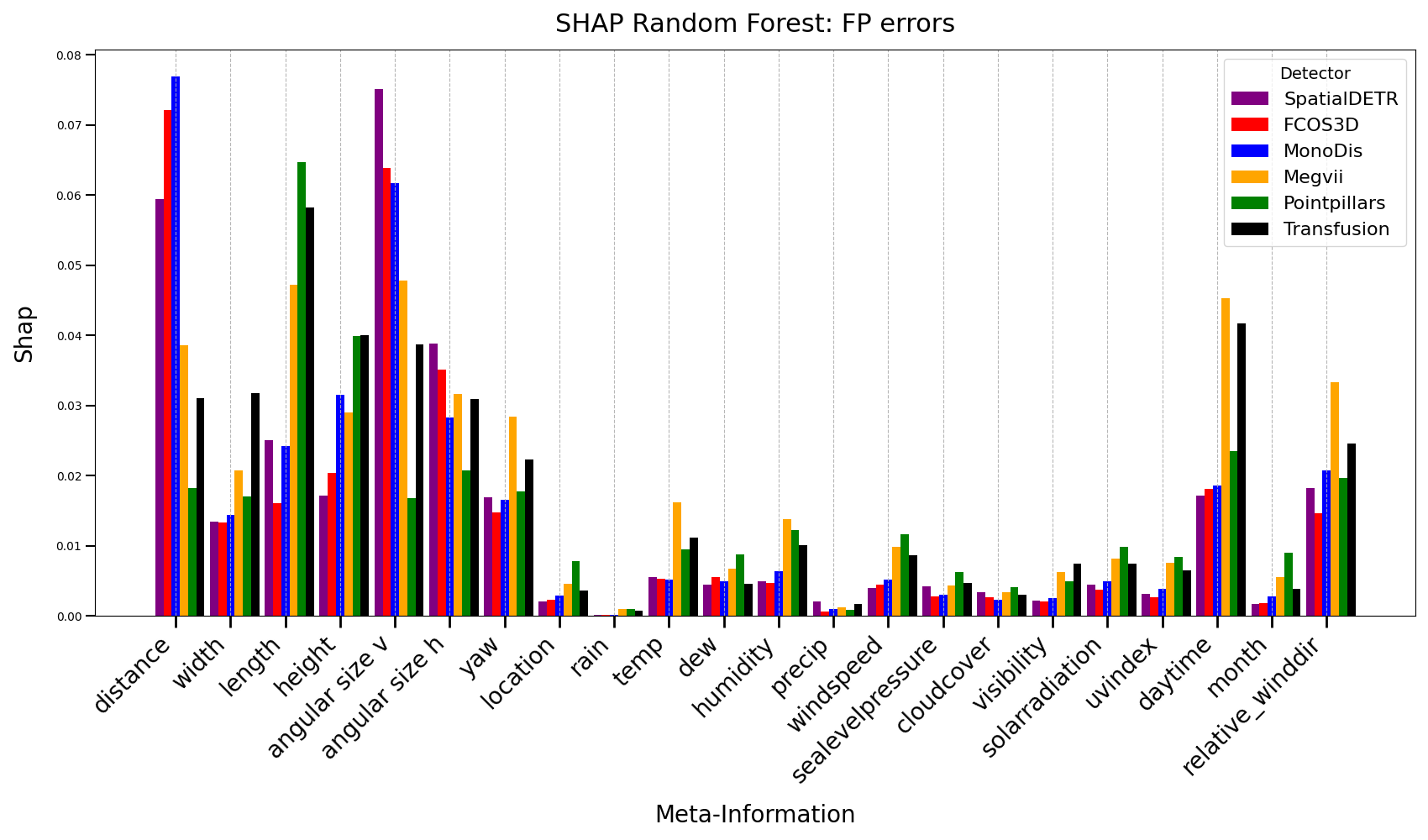}
   \caption{A barplot of the mean absolute shap values between the FP object detection error on pedestrians and the meta-information}
   \label{fig:Shap_FP}
  \end{minipage}
\end{figure*}

This section analyzes the detection performance based on whether the predicted bounding box is a TP or an FP, i.e. ghost detection. Except for PointPillars, the LiDAR-based detectors show better performance in terms of precision compared to the camera detectors, as illustrated in Table \ref{tbl:recall}. Fig. \ref{fig:Kendall_TPFP} shows the bar plot of Kendall's Tau correlation coefficient, which illustrates the relationship between detection performance in terms of FP or TP and the meta-information. In contrast to the previous analysis, the FP case lacks ground-truth information. Consequently, the object-specific meta-information is obtained from the predicted bounding box generated by the detector. As can be observed in Fig. \ref{fig:Kendall_TPFP}, the correlation coefficients of the three camera-based detectors have a larger negative value for the distance and a larger positive value for the angular sizes of the predicted bounding boxes compared to the LiDAR detectors. This suggests that the number of pedestrian ghost detections by the camera-based detectors increases with distance. Furthermore, the monotonic dependence of weather, time, and location on the detection performance is generally low for all detectors except for PointPillars. PointPillars shows stronger positive dependencies with increasing temperatures and in the One-North district in Singapore, as well as some stronger negative correlations with month and humidity, compared to the other detectors. Nevertheless, coefficients remain relatively small in total. Variations in the relationship between detector performance and predicted bounding box sizes are also observed across different detectors, regardless of modality. Fig. \ref{fig:MI_FPTP} illustrates the MI value between the meta-information and the FP Errors. As previously observed in Kendall's Tau correlation analysis shown in Fig. \ref{fig:Kendall_TPFP}, distance and angular sizes are more influential on the camera-based detectors than on the LiDAR-based detectors. Aside from precipitation and the sea level pressure, the LiDAR-based detectors are generally more influenced by weather and location. In particular, PointPillars tends to have higher MI values, while Transfusion-L tends to have lower values. In addition, wind direction and time of day have the greatest impact on the performance of the LiDAR detectors. Additionally, among the camera-based detectors, MonoDis is generally more influenced by environmental conditions. Finally, the size and heading of the predicted bounding box have a smaller impact on the FP errors of the detectors. Next, the meta-model-based analysis is examined. Fig. \ref{fig:Shap_FP} shows the mean absolute SHAP values of the meta-information used to train the RF model that predicts whether the output of the object detector is TP or FP. Similar to the univariate analysis, distance and vertical angular size have a stronger influence on the performance of camera-based detectors compared to LiDAR-based detectors. However, the horizontal angular size has a smaller effect on camera-based detectors and is also comparable to its effect on LiDAR detectors, contrary to the MI analysis in Fig. \ref{fig:MI_FPTP} and the Kendall's Tau analysis in Fig. \ref{fig:Kendall_TPFP}. In contrast to the univariate analysis, the predicted bounding box length is the most influential factor for the LiDAR detectors. The meta-model-based analysis further confirms that meta-information related to weather and location has a comparatively small impact, while predicted bounding box parameters play a more significant role. Nevertheless, the mean absolute SHAP values in Fig. \ref{fig:MI_FPTP} for temperature and humidity are slightly higher than for the camera-based detectors. Apart from the differences observed between the two sensor modalities, differences between the different detectors can be observed, especially between the LiDAR-based detectors, as in the MI analysis. While the LiDAR-based detectors Megvii and Transfusion-L generally show similar trends, PointPillars often behaves differently, for instance, daytime strongly affects Megvii and Transfusion-L, while PointPillars is less affected, similar to the camera detectors.

 \section{Discussion}
In our analysis of the factors influencing the performance of single-sensor object detection 3D pedestrian detection, we identified several key findings. Camera-based detectors tend to misdetect distant pedestrians more frequently and generate more FP detections at greater distances from the ego-vehicle compared to the LiDAR-based detectors. This may be due to the fact that the LiDAR sensor in NuScenes may have a larger detection range than the corresponding cameras. In addition, FN errors are affected by the occlusion of pedestrians for all detectors examined, with a stronger effect on camera-based detectors. In contrast, LiDAR-based detectors are more dependent on the pedestrian's attribute and the pedestrian's velocity, which is naturally related to the attribute. In particular, pedestrians who are standing, sitting, or lying down are more likely to be misdetected by the LiDAR-based detectors than moving pedestrians. Torres et al. \cite{Torres2023LidarPed} found that the number of reflections and intensities in 3D point clouds vary with human pose. This could complicate the detection performance of LiDAR detectors, as observed in our analysis. Furthermore, our analysis shows that LiDAR-based detectors are more sensitive to meta-information related to weather conditions than camera detectors, especially with respect to FP errors. However, the overall influence remains relatively small, despite the direct impact of adverse weather conditions on sensor signal quality. The weather-related meta-information in NuScenes is extracted from the nearest weather station to the ego-vehicle. Aside from the fact that certain factors, such as precipitation, cloud cover, or wind speed can vary locally within a few kilometers, the main potential problem is the limited diversity of weather in the dataset. First, the recorded weather information in NuScenes remains unchanged within each 20-second scene sequence. The examined dataset consists of only 150 scenes, while it contains about 6000 different samples. In addition, the dataset lacks sufficient variability in environmental conditions. The lowest recorded value of visibility is $6.3 \si{\kilo\meter}$, while haze and fog typically occur at visibilites less than $4 \si{\kilo\meter}$ and $0.5 \si{\kilo\meter}$ \cite{visibilitySubekti2019}, respectively. Similarly, the lowest recorded temperature is $ 21.7 \si{\degreeCelsius}$, indicating that all recordings were made under warm weather conditions. Lastly, the precipitation does not exceed $3.4 \si{\frac{\milli\meter}{\hour}}$, while heavy rain is generally classified as exceeding $6 \si{\frac{\milli\meter}{\hour}}$ \cite{Avanzato2020}. In addition, consistent with our findings and the work of Ponn et al. \cite{Ponn.2020}, the rain annotation provided for each sample in NuScenes has a minimal impact on detection performance, further indicating that the dataset contains only low intensity rain conditions. \\
Our analysis also reveals distinct differences between the object detectors examined, particularly for the LiDAR-based detector, where PointPillars is substantially different. Furthermore, the meta-model-based analysis shows some differences compared to the univariate analysis. In particular, the FN error analysis for PointPillars in the meta-model-based analysis in Fig. \ref{fig:Shap_FN} differs strongly from the univariate analysis shown in Fig. \ref{fig:Kendall_TPFN} and Fig. \ref{fig:MI_TPFN}. This discrepancy suggests that the influence of certain factors varies when combined with other meta-information. The results of the metamodel-based analysis do not necessarily reflect the true behavior of the object detectors, even though the trained RF takes into account correlations between meta-information and that SHAP values are computed via conditional sampling to avoid extrapolation. This limitation arises from the so-called Rashomon Effect in ML \cite{freiesleben2024supervised}, which describes how multiple ML models that perform similarly on a given task can have structurally different decision boundaries. Consequently, the interpretations derived from the meta-model-based approach do not necessarily represent causal relationships.   

\section{Conclusion \& Outlook}
In our work, we developed a methodology to identify the influence of various environmental and object-specific factors from the object to be detected on the performance of 3D object detectors. Our analysis is based on the 3D detection of pedestrians, which is a highly safety-relevant class of objects in AD. By examining the characteristics of the data, our approach allows a comparative assessment of the strength and direction of influence of different factors, while identifying potential sources of error. Our developed methodology helps to identify performance insufficiencies of the object detectors and thus supports the development of safe DNN-based object detectors in AD. In addition, the RF trained on detecting FP errors could serve as the basis for a monitor to prevent false predictions of the deployed object detector. However, the extraction of potential error sources is limited by the meta-information contained in NuScenes \cite{Caesar.2020}, especially the low diversity of adverse weather conditions. Future research could extend the analysis to multivariate causal methods to enhance the investigation of error sources and limitations in specific 3D object detectors. This approach could allow the identification of challenging scenarios characterized by a combination of influencing factors, which is planned for future work. Furthermore, the analysis of perception errors on other datasets could help to validate our findings. Finally, our approach could be further extended by using explainable AI in the form of interpretable surrogate models or auxiliary explanations \cite{Kuznietsov2024XAI}, which could improve the identification of error sources in object detection. 
\bibliographystyle{IEEEtran}
\bibliography{failure_causes.bib}

\end{document}